\pdfoutput=1

\documentclass[11pt]{article}

\usepackage{acl}

\usepackage{times}
\usepackage{latexsym}

\usepackage[T1]{fontenc}

\usepackage[utf8]{inputenc}

\usepackage{microtype}

\usepackage{amsmath}
\usepackage{xcolor}%
\usepackage{upquote}

\usepackage{arydshln}
\usepackage{easyReview}

\newcommand{\bart}{\textsc{Bart}}

\usepackage{bibentry}

\usepackage{times}
\usepackage{latexsym}

\usepackage{booktabs}

\usepackage{textcomp}
\usepackage{graphicx}
\usepackage{multirow} 
\usepackage{enumitem}

\usepackage{authblk}

\newcommand{\oneS}{\ensuremath{{}^{\textstyle *}} }

\title{\textsc{Tstr}: Too Short to Represent, Summarize with Details! \\Intro-Guided Extended Summary Generation}

\author[ ]{\textbf{Sajad Sotudeh}}
\author[ ]{\textbf{Nazli Goharian}}
\affil[ ]{IR Lab, Georgetown University, Washington DC 20057, USA}
\affil[ ]{\normalsize \texttt {\{sajad, nazli\}@ir.cs.georgetown.edu}}

\begin{document}
\maketitle
\begin{abstract}
Many scientific papers such as those in arXiv and PubMed data collections have abstracts with varying lengths of 50--1000 {words} and average length of approximately 200 {words}, where longer abstracts typically convey more information about the source paper. Up to recently, scientific summarization research has typically focused on generating short, abstract-like summaries following the existing datasets used for scientific summarization. In domains where the source text is relatively long-form, such as in scientific documents, such summary is not able to go beyond the general and coarse overview and provide salient information from the source document. The recent interest to tackle this problem motivated curation of scientific datasets, arXiv-Long and PubMed-Long, containing human-written summaries of 400-600 {words}, hence, providing a venue for research in generating long/extended summaries. Extended summaries facilitate a faster read while providing details beyond coarse information. In this paper, we propose \textsc{Tstr}, an extractive summarizer that utilizes the introductory information of documents as pointers to their salient information. The evaluations on two existing large-scale extended summarization datasets indicate statistically significant improvement in terms of \textsc{Rouge} and average \textsc{Rouge (F1)} scores (except in one case) as compared to strong baselines and state-of-the-art. Comprehensive human evaluations favor our generated extended summaries in terms of cohesion and completeness.
\end{abstract}

\section{Introduction}
\label{sec:intro}

\begin{figure}[t]
    \centering
    \includegraphics[scale=0.5]{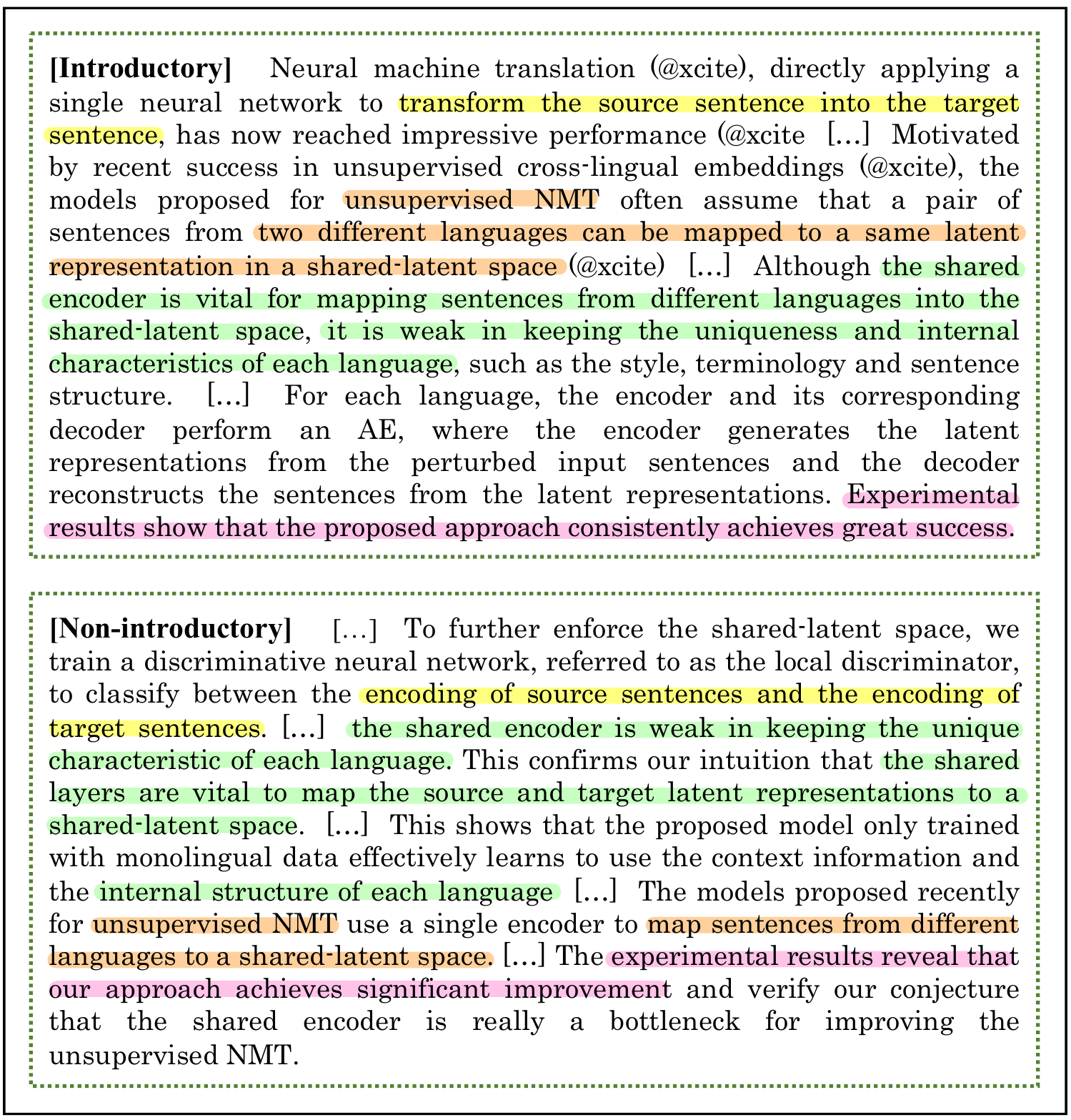}
    \caption{A truncated human-written extended summary. Top box: introductory information, bottom box: non-introductory information. Colored spans are pointers from introductory sentences to associated non-introductory detailed sentences.}
    \label{fig:my_label}
\end{figure}

Over the past few years, summarization task has witnessed a huge deal of progress in extractive~\cite{Nallapati2017SummaRuNNerAR, Liu2019TextSW, Yuan2020FactlevelES, Cui2020EnhancingET, Jia2020DistilSumDT, Feng2018AttentiveEE} and abstractive~\cite{See2017GetTT, Cohan2018ADA, Gehrmann2018BottomUpAS, zhang2019pegasus, Tian2019AspectAO, Zou2020PretrainingFA} settings. Many scientific papers such as those in arXiv and PubMed~\cite{Cohan2018ADA} posses abstracts of varying length, ranging from 50 to 1000 {words} and average length of approximately 200 {words}. While scientific paper summarization has been an active research area, most works~\cite{Cohan2018ADA, Xiao2019ExtractiveSO, Cui2021SlidingSN, Rohde2021HierarchicalLF} in this domain have focused on generating typical short and abstract-like summaries~\cite{Chandrasekaran2020LongSumm}.
Short summaries might be adequate when the source text is of short-form such as those in news domain; however, to summarize longer documents such as scientific papers, an extended summary including 400--600 terms on average, such as those found in extended summarization datasets of arXiv-Long and PubMed-Long, is more appealing as it conveys more detailed information. 

Extended summary generation has been of research interest very recently. \citet{Chandrasekaran2020LongSumm} motivated the necessity of generating extended summaries through LongSumm shared task~\footnote{\url{https://ornlcda.github.io/SDProc/sharedtasks.html}}. Long documents such as scientific papers are usually framed in a specific structure. They start by presenting general \textit{introductory information}~\footnote{We will exchangeably use \textsl{(non-)introductory information} and \textsl{(non-)introductory sentences} in the rest of this paper.}. This introductory information is then followed by supplemental information (i.e., non-introductory) that explain the 
initial introductory information in more detail. Similarly, as shown in Figure 1, this pattern holds in a human-written extended summary of a long document, where the preceding sentences (top box inside Figure \ref{fig:my_label}) are introductory sentences and succeeding sentences (bottom box inside Figure \ref{fig:my_label}) are explanations of the introductory sentences. In this study, we aim to guide the summarization model to utilize the aforementioned rationale in human-written summaries. 
\textcolor{black}{We consider \textbf{introductory sentences} as those that appear in the first section of paper with headings such as \textit{Introduction}, \textit{Overview}, \textit{Motivations}, and so forth. As such, all other parts of paper and their sentences are considered as \textbf{non-introductory} (i.e., \textbf{supplementary}). We use these definitions in the reminder of this paper}.


Herein, we approach the problem of \textit{extended} summary generation by incorporating the most important introductory information into the summarization model.  We hypothesize that incorporating such information into the summarization model guides the model to pick salient detailed non-introductory information to augment the final extended summary. \textcolor{black}{The importance of the role of introduction in the scientific papers was earlier presented in
~\cite{Teufel2002SummarizingSA,Armaan2013HowTW,Jirge2017PreparingAP} where they showed such information provides clues (i.e. pointers) to the objectives and experiments of studies. 
Similarly, \citet{Boni2020ASO} conducted a study to show the importance of introduction part of scientific papers as its relevance to the paper's abstract.}
To validate our hypothesis, we test the proposed approach on two publicly available large-scale extended summarization datasets, namely arXiv-Long and PubMed-Long. Our experimental results improve over the strong baselines and state-of-the-art models. In short, the contributions of this work are as follows:
\begin{itemize}
    \item A novel multi-tasking approach that incorporates the salient introductory information into the extractive summarizer to guide the model in generating a 600-term (roughly) extended summary of a long document, containing the key detailed information of a scientific paper.
    
    \item Intrinsic evaluation that demonstrates statistically significant improvements over strong extractive and abstractive summarization baselines and state-of-the-art models.
    
    \item An extensive human evaluation which reveals the advantage of the proposed model in terms of cohesion and completeness.

\end{itemize}

\section{Related Work}



Summarizing scientific documents has gained a huge deal of attention from researchers, although it has been studied for decades. Neural efforts in scientific text have used specific characteristics of papers such as discourse structure \cite{Cohan2018ADA, Xiao2019ExtractiveSO} and citation information~\cite{Qazvinian2008ScientificPS, Cohan2015ScientificAS, Cohan2018-2} to aid summarization model. 
While prior work has mostly covered the generation of shorter-form summaries (approx. 200 terms), generating extended summaries of roughly 600 terms for long-form source documents such as scientific papers has been motivated very recently~\cite{Chandrasekaran2020LongSumm}.

The proposed models for the extended summary generation task include jointly learning to predict sentence importance and sentence section to extract top sentences \cite{sotudeh-gharebagh-etal-2020-guir}; utilizing section-contribution computations to pick sentences from important section for forming the final summary~\cite{ghosh-roy-etal-2020-summaformers}; identifying salient sections for generating abstractive summaries~\cite{gidiotis-etal-2020-auth};  ensembling of extraction and abstraction models to form final summary~\cite{ying-etal-2021-longsumm}; an extractive model with TextRank algorithm equipped with BM25 as similarity function~\cite{kaushik-etal-2021-cnlp};  and incorporating sentences embeddings into graph-based  extractive summarizer in an unsupervised manner~\cite{RamirezOrta2021UnsupervisedDS}. Unlike these works, we do not exploit any sectional nor citation information in this work. To the best of our knowledge, we are the first at proposing the novel method of utilizing introductory information of the scientific paper to guide the model to learn to generate summary from the salient and related information. 

\section{{Background: Contextualized language models for summarization}}
\label{sec:approach}


\label{sect:bg}
Contextualized language models such as \textsc{Bert}~\cite{Devlin2019BERTPO}, and \textsc{RoBerta}~\cite{Liu2019RoBERTaAR} have achieved state-of-the-art performance on a variety of downstream NLP tasks including text summarization. \citet{Liu2019TextSW} were the first to fine-tune a contextualized language model (i.e., \textsc{Bert}) for the summarization task. \textcolor{black}{They proposed \textsc{BertSum} ---a fine-tuning scheme for text summarization--- that outputs the sentence representations of the 
source
document (we use the term source and source document interchangeably, referring to the entire document).} \textcolor{black}{The \textsc{BertSumExt} model, which is built based on \textsc{BertSum}, was proposed for the extractive summarization task. It utilizes the representations produced by \textsc{BertSum}, passes them through Transformers encoder~\cite{Vaswani2017Att}, and finally uses a linear layer with Sigmoid function to compute copying probabilities for each input sentence.} Formally, let  $l_1, l_2, ..., l_n$  be the binary tags over the source sentences $\mathbf{x}=\mathbf{\{}sent_1, sent_2, ..., sent_n\mathbf{\}}$ of a long document, in which $n$ is the number of sentences in the paper. The \textsc{BertSumExt} network runs over the source documents as follows (Eq. \ref{eq1}),

\vspace{0.1em}
\begin{equation}
    \begin{array}{l}
   
    {h_b} = \mathbf{BertSum}({\mathbf{x}}) \vspace{0.2em}\\\vspace{0.2em}
    {h} = \mathbf{Encoder_{t}}({h_b})\vspace{0.2em}\\\vspace{0.2em}
     {{p} = {\sigma}{(W_{o} h + b_o)}}
    
  \end{array}
  \label{eq1}
\end{equation}
where $\mathbf{h_b}$ and $\mathbf{h}$ are the representations of source sentences encoded by \textsc{BertSum} and Trasformers encoder, respectively. $W_o$ and $b_o$ are trainable parameters, and
${p}$ is the probability distribution over the source sentences, signifying extraction copy likelihood. The goal of this network is to train a network that can identify the positive sets of sentences as the summary. To prevent the network from selecting redundant sentences, \textsc{BertSum} uses \textit{Trigram Blocking}~\cite{Liu2019TextSW} for sentence selection in inference time. We refer the reader to the main paper for more details. 

\begin{figure}[t]
    \centering
    \includegraphics[scale=0.22]{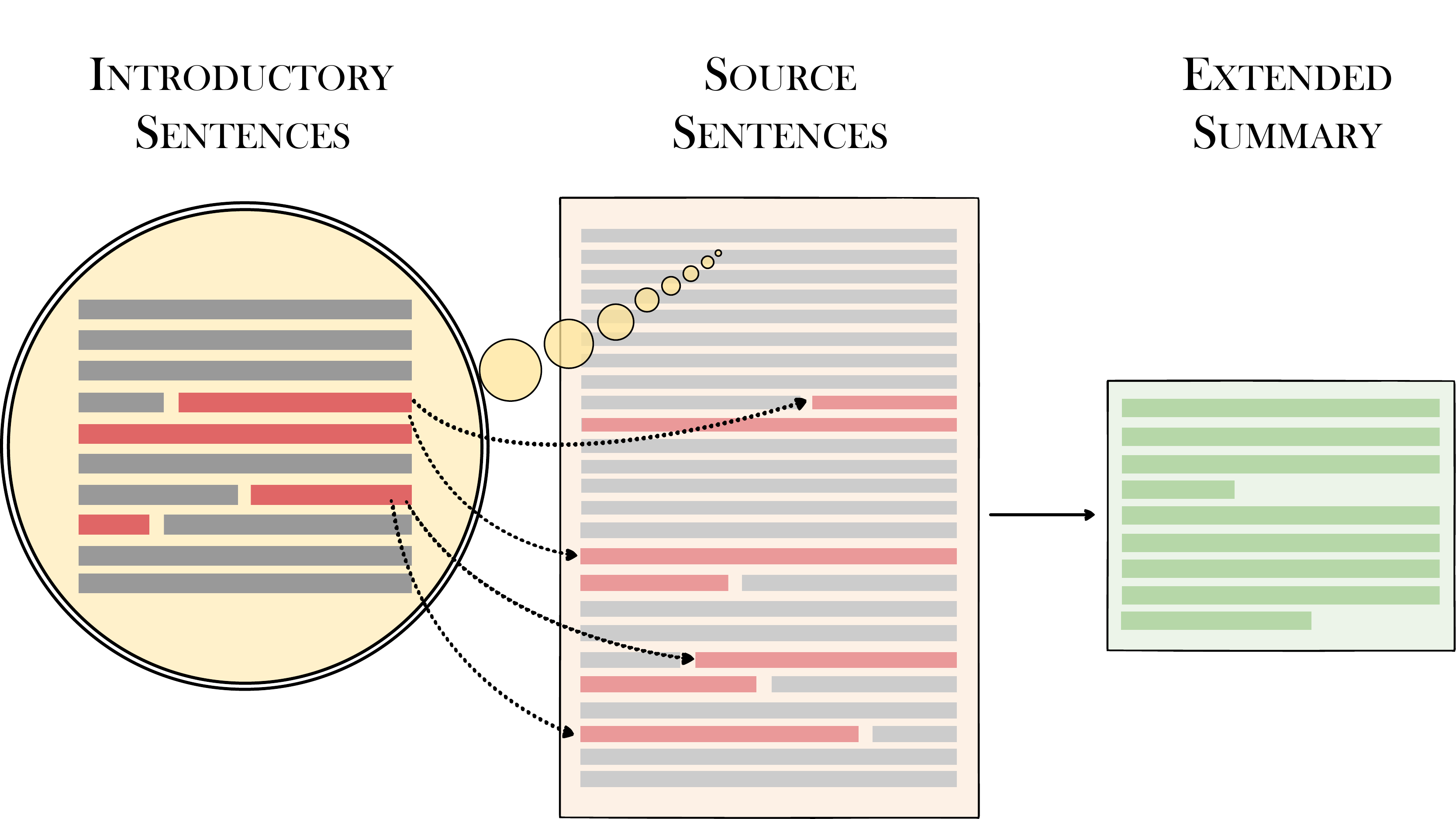}
    \caption{Our model uses introductory sentences as pointers to the source sentences. It then forms the final extended summary by extracting salient sentences from the source. Highlights in red show the salient parts.}
    \label{fig:model_overview}
\end{figure}

\begin{figure*}
\centering
\includegraphics[scale=0.27]{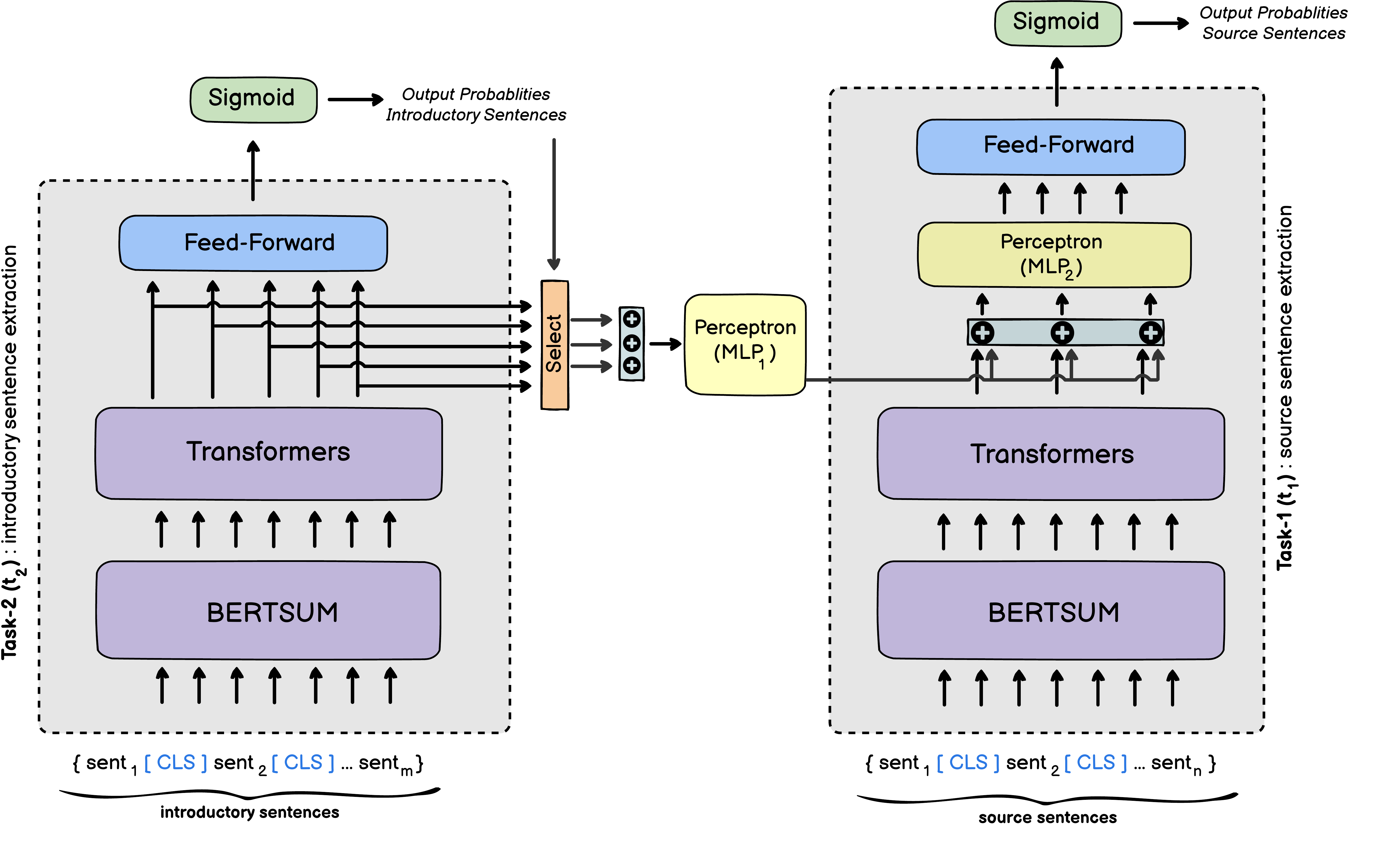}
\caption{Detailed illustration of our summarization framework. \textbf{Task-1 ($\mathbf{t_1}$):} source sentence extraction (right-hand gray box). \textbf{Task-2 ($\mathbf{t_2}$):} introductory sentence extraction (left-hand gray box). As shown, the identified salient introductory sentences at training stages are incorporated into the representations of source sentences by the $\texttt{\textbf{Select}}(\cdot)$ function (orange box) with ${k}=3$. Plus sign shows the concatenation layer. The feed-forward neural network is made of one linear layer. 
}
\label{fig:model}
\label{fig:summ_model}
\end{figure*}

\section{{\textsc{Tstr}:} Intro-guided Summarization}
\label{sect:our_model}

 In this {section},  we describe our methodology to tackle the extended summary generation task. Our approach exploits the introductory information~\footnote{Introductory information is defined in Section \ref{sec:intro}}. of the paper as pointers to salient sentences within it, as shown in Figure \ref{fig:model_overview}. It is ultimately expected that the extractive summarizer is guided to pick salient sentences across the entire paper.

The detailed illustration of our model is shown in Figure \ref{fig:model}. To aid the extractive summarization model (i.e., right-hand box in Figure \ref{fig:model}) which takes in source sentences of a scientific paper, we utilize an additional \textsc{BertSum} encoder called Introductory encoder (left-hand box in Fig. \ref{fig:model}) that receives $\mathbf{x_{intro}} = \mathbf{\{}sent_1, sent_2, ..., sent_m\mathbf{\}}$, with $m$ being the number of sentences in introductory section. \textcolor{black}{The aim of adding second encoder in this framework is to identify the clues in the introductory section which point to the salient supplementary sentences~\footnote{Supplementary sentences are defined in Section \ref{sec:intro}.}.} The $\textsc{BertSum}$ network computes the extraction probabilities for {introductory} sentences as follow (same way as in Eq. \ref{eq1}),

\begin{equation}
    \begin{array}{l}
         {\tilde{h}_b} = \mathbf{BertSum(x_{intro})} \vspace{0.2em}\\ \vspace{0.2em}
         {\tilde{h}} = \mathbf{Encoder_{t}}(\tilde{h}_b) \vspace{0.2em}\\\vspace{0.2em}
          {{\tilde{p}} = {\sigma}{(W_{j} \tilde{h} + b_j)}}
    \end{array}
    \label{eq:intro}
\end{equation}
in which $\tilde{h}_b$, and $\tilde{h}$ are the introductory sentence representations by \textsc{BertSum}, Transformers encoder, respectively. ${\tilde{p}}$ is the introductory sentence extraction probabilities. $W_j$ and $b_j$ are trainable matrices. 

After identifying salient introductory sentences, the representations associated with them are retrieved using a pooling function and further used to guide the first task (i.e., right-hand side in Figure \ref{fig:model}) as follows, 
\begin{equation}
    \begin{array}{l} 
         {\tilde{h}_{top}} =  \texttt{\textbf{Select}} (\tilde{h}, {\tilde{p}}, k) 
         \vspace{0.5em}
         \\ 
         \vspace{0.1em}
         
         {\hat{h}} =  \mathbf{MLP}_{\mathbf{1}}({\tilde{h}_{top}}) 
    \end{array}
    \label{eq:intro}
\end{equation}
 {where $\texttt{\textbf{Select}}(\cdot)$ is a function that takes in all introductory sentence representations (i.e., $\tilde{h}$), and introductory sentence probabilities $\tilde{p}$. It then outputs the representations associated with top ${k}$ introductory sentences, sorted by $\tilde{p}$. To extract top introductory sentences, we first sort $\tilde{h}$ vectors based on their computed probabilities $\tilde{p}$ and then we pick up top $k$ hidden vectors (i.e., $\tilde{h}_{top}$) that has the highest probability.  $\mathbf{MLP}_\mathbf{1}$ is a multi-layer perceptron that takes in concatenated vector of top introductory sentences and projects it into a new vector called $\hat{h}$. }
 
At the final stage, we concatenate {the transformed} introductory top sentence representations (i.e., $\hat{h}$) with each source sentence representations from Eq. \ref{eq1} (i.e., $h_i$ where $i$ shows the $i$th paper sentence) {and process them to produce a resulting vector ${r}$ which is \textit{intro-aware} source sentence hidden representations.} After processing the resulting vector through a linear output layer (with $W_z$ and $b_z$ as trainable parameters), we obtain final {\textit{intro-aware}} sentence extraction probabilities (i.e., ${{p}}$) as follows,
 \vspace{0.9em}
\begin{equation}
\begin{array}{l}
    {r} = \mathbf{MLP}_{\mathbf{2}}(h_i \hspace{0.2em} ; \hspace{0.2em} {\hat{h}}) \vspace{0.5em} \\ \vspace{0.3em}
     {{{p}} = {\sigma}{(W_{z} {r} + b_z)}}


\end{array}
\end{equation}
in which $\mathbf{MLP_2}$ is a multi-layer perceptron, {influencing the knowledge from introductory sentence extraction task (i.e., $\mathbf{t_2})$ into the source sentence extraction task (i.e., $\mathbf{t_1}$).} We train both tasks through our end-to-end system jointly as follows, 

\begin{equation}
    \ell_{\mathbf{{total}}} = (\alpha) \ell_{\mathbf{t_1}} + (1-\alpha) \ell_{\mathbf{t_2}}
    \label{eq:mlt}
\end{equation}
where $\ell_\mathbf{t_1}$, and $\ell_\mathbf{t_2}$ are the losses computed for introductory sentence extraction and source sentence extraction tasks, $\alpha$ is the regularizing parameter that balances the learning process between two tasks, and $\ell_\mathbf{total}$ is the total computed loss that is optimized during the training.

\section{Experimental Setup}
In this section, we explain the datasets, baselines, and preprocessing and training parameters. 

\subsection{Dataset}
We use two publicly available scientific extended summarization datasets~\cite{Sotudeh2020OnGE}.

\begin{itemize}[leftmargin=*,label={-}]
    
\item \textbf{arXiv-Long: }  A  set of arXiv scientific papers containing papers from various scientific domains such as physics, mathematics, computer science, quantitative biology. arXiv-Long is intended for extended summarization task and was filtered from a larger dataset i.e., arXiv \cite{Cohan2018ADA} for the summaries of more than 350 tokens. The ground-truth summaries (i.e., abstract) are long, with the average length of 574 tokens. It contains 7816 (train), 1381 (validation), and 1952 (test) papers.



\item\textbf{PubMed-Long: } A set of biomedical scientific papers from PubMed with average summary length of 403 tokens. This dataset contains 79893 (train), 4406 (validation), and 4402 (test) scientific papers.

\item \textbf{LongSumm: } The recently proposed {LongSumm} dataset for a shared task ~\cite{Chandrasekaran2020LongSumm} contains 2236 abstractive and extractive summaries for training and 22 papers for the official test set. We report a comparison with \textsc{BertSumExtMulti} using this data in Table \ref{tab:blind}. However, as the official test set is blind, our experimental results in Table \ref{tab:main} do not use this dataset.
\end{itemize}

\subsection{Baselines}
We compare our model with two strong non-neural systems, and four state-of-the-art neural summarizers.
We use all of these baselines for the purpose of extended summary generation whose documents hold different characteristics in length, writing style, and discourse structure as compared to documents in the other domains of summarization.

\begin{itemize}[leftmargin=*,label={-}]


\item \textbf{\textsc{LSA}}~\cite{Steinberger2004LSA}: an extractive vector-based model that utilizes Singular Value Decomposition (SVD) to find the semantically important sentences.


\item \textbf{\textsc{LexRank}}~\cite{Erkan2004LexRankGL}: 
a widely adopted extractive summarization baseline that utilizes a graph-based approach based on eigenvector centrality to identify the most salient sentences. 


\item \textbf{\textsc{BertSumExt}}~\cite{Liu2019TextSW}: a contextualized summarizer fine-tuned for summarization task, which encodes input sentence representations, and then processes them through a multi-layer Transformers encoder to obtain document-level sentence representation. Finally, a linear output layer with Sigmoid activation function outputs a probability distribution over each input sentence, denoting the extent to which they are probable to be extracted. 


\textcolor{black}{\item \textbf{\textsc{BertSumExt-Intro}}~\cite{Liu2019TextSW}: a \textsc{BertSumExt} model that only runs on the introductory sentences as the input, and extracts the salient introductory sentences as the summary. 
}


\item \textbf{\textsc{BertSumExtMulti}}~\cite{Sotudeh2020OnGE}: an extension of the  \textsc{BertSumExt} model that incorporates an additional linear layer with Sigmoid classifier to output a probability distribution over a fixed number of pre-defined sections that an input sentence might belong to.  The additional network is expected to predict a single section for an input sentence and is trained jointly with \textsc{BertSumExt} module (i.e., sentence extractor). 


\item \textbf{\textsc{Bart}}~\cite{Lewis2020BARTDS}: a state-of-the-art abstractive summarization model that makes use of pretrained encoder and decoder. \textsc{Bart} can be thought of as an extension of \textsc{BertSum} in which merely encoder is pre-trained, but decoder is trained from scratch. \textcolor{black}{While our model is an extractive one, at the same time, we find it of value to measure the abstractive model performance in the extended summary generation task.}


\end{itemize}

\subsection{Preprocessing, parameters, labeling, and implementation details}
We used the open implementation of  \textsc{BertSumExt} with default parameters~\footnote{\url{https://github.com/nlpyang/PreSumm}}. To implement the non-neural baseline models, we utilized Sumy python package~\footnote{\url{https://github.com/miso-belica/sumy}}. Longformer model ~\cite{Beltagy2020LongformerTL} is utilized as our contextualized language model for running all the models {due to its efficacy at processing long documents.} For our model, the cross-entropy loss function is set for two tasks (i.e.,  $\mathbf{t_1}:$ source sentence extraction and $\mathbf{t_2}:$ introductory sentences extraction in Figure \ref{fig:model}) and the model is optimized through multi-tasking approach as discussed in Section \ref{sec:approach}. The model with the highest \textsc{Rouge-2} on validation set is selected for inference. The validation is performed every 2k training steps. $\alpha$ (in Eq. \ref{eq:mlt}) is set to be 0.5 \textcolor{black}{(empirically determined)}. Our model includes 474M trainable parameters, trained on dual GeForce GTX 1080Ti GPUs for approximately a week.  We use ${k}=5$ for arXiv-Long, ${k}=8$ for PubMed-Long datasets  (Eq. \ref{eq:intro}). 
We make our model implementation as well as sample summaries publicly available to expedite ongoing research in this direction~\footnote{\url{https://github.com/Georgetown-IR-Lab/TSTRSum}}.

A two-stage labeling approach was employed to identify ground-truth introductory and non-introductory sentences. In the first stage, we used a greedy labeling approach~\cite{Liu2019TextSW} to label sentences within the first section of a given paper
(i.e., labeling introductory sentences)  with respect to their \textsc{Rouge} overlap~\footnote{We used mean of \textsc{Rouge-2} and \textsc{Rouge-L}.} with the ground-truth summary (i.e., abstract). In the second stage, the same greedy approach was exploited over the rest of sentences (i.e., non-introductory)\footnote{We assumed that non-introductory sentences occur in sections other than the first section.} with regard to their \textsc{Rouge} overlap with the identified introductory sentences in the first stage. Our choice of \textsc{Rouge-2} and \textsc{Rouge-L} is based on the fact that these express higher similarity with human judgments~\cite{Cohan2016RevisitingSE}. We continued the second stage until a fixed length of the summary was reached. Specifically, the fixed length of positive labels is set to be 15 for arXiv-Long, and 20 for PubMed-Long datasets as these achieved the highest oracle \textsc{Rouge} scores in our experiments.





\begin{table*}[t]
\centering 
\begin{center}


\scalebox{1}{
\begin{tabular*}{\textwidth}{llllllllll}
\toprule

      & \multicolumn{4}{c}{\textit{arXiv-Long}} &   & \multicolumn{4}{c}{\textit{PubMed-Long}} \\
 \cline{2-5} \cline{7-10}
 
 Model                    & \small R1(\%)  &\small R2(\%)  &\small RL(\%) & \small F1 (\%)  & &\small R1(\%)  &\small R2(\%)  &\small RL(\%) & \small F1 (\%) \\

 \midrule
 
 \textsc{Oracle}      &
  {53.35} &	 {24.40} &
  {23.65}  &
  {33.80}
& &

  {52.11}&	{23.41}&	 {25.42} 
  & {33.65}\\
   \textsc{BertSumExt-Intro}      &
  {44.88} &	 {15.99} &
  {19.14}  &
  {26.25}
& &

  {45.08}&	{20.08}&	 {21.52} 
  & {28.89}\\
 \vspace{-1em} \\
    \hdashline[0.5pt/2pt] \vspace{-1em} \\
  \textsc{LSA}                &  {43.23} &	 {13.47} &	 {17.50} &  {24.73}   & &  {44.47}&	 {15.38}&	 {19.17} &   {26.34}\\

  \textsc{LexRank}                   &  {43.73} &	 {15.01} &	 {18.62} &  {25.41}  & &  {48.63}&	 {20.37}&	 {22.49} &  {30.50}\\

 \textsc{BertSumExt}&  {48.42} &	 {19.71} &	 {21.47} &  {29.87} & &   {48.82}&	 {20.89}&	 {23.37} &  {31.03} \\

  \textsc{BertSumExtMulti}                    &  {48.52} &	 {19.66} &	 {21.42} &  {29.87} & &  {48.85}&	 {20.71}&	 {23.29} &  {30.95}\\

  \textsc{Bart}               &  {48.12} &	 {15.30} &	 {20.80} &  {28.07}            &     &  {48.32} &	 {17.33} &	 {21.42} &  {29.87}\\

  \textsc{Tstr} (Ours)     \hspace{1em}  &
  \textbf{49.20\oneS} &	 \textbf{20.19\oneS} &
  \textbf{22.22\oneS}  &
  {\textbf{30.54}}
& &

  \textbf{49.32\oneS}&	 \textbf{21.41\oneS}&	 \textbf{23.67} 
  & \textbf{31.47}\\
  
\bottomrule
\end{tabular*}
}
\end{center}


\caption{\textsc{Rouge (F1)} results of the baseline models and our model on the test sets of the extended summarization datasets (arXiv-Long, and PubMed-Long). \oneS shows the statistical significance (paired t-test, $p<0.05$). }
\label{tab:main}

\end{table*}

\section{Results}
\begin{table}

\centering 
\begin{center}
\centering 
\resizebox{\columnwidth}{!}{%
 \begin{tabular}{ lrrrc}
 \toprule
  & \fontsize{10}{60}\selectfont R1 & \fontsize{10}{60}\selectfont R2  & \fontsize{10}{60}\selectfont RL & F1(\%) \\
 \midrule

\fontsize{10.5}{60}\selectfont Summaformers~\shortcite{ghosh-roy-etal-2020-summaformers} & \fontsize{11}{60}\selectfont 49.38 & \fontsize{11}{60}\selectfont \textbf{16.86} & \fontsize{11}{60}\selectfont \textbf{21.38} &  \fontsize{11}{60}\selectfont 29.21  \\

\fontsize{10.5}{60}\selectfont IIITBH-IITP~\citeyearpar{reddy-etal-2020-iiitbh} & \fontsize{11}{60}\selectfont 49.03 & \fontsize{11}{60}\selectfont 15.74 & \fontsize{11}{60}\selectfont 20.46 &  \fontsize{11}{60}\selectfont 28.41 \\

\fontsize{10.5}{60}\selectfont Auth-Team~\citeyearpar{gidiotis-etal-2020-auth} & \fontsize{11}{60}\selectfont 50.11 & \fontsize{11}{60}\selectfont 15.37 & \fontsize{11}{60}\selectfont 19.59 &  \fontsize{11}{60}\selectfont 28.36 \\

\fontsize{10.5}{60}\selectfont CIST\_BUPT~\citeyearpar{li-etal-2020-cist} & \fontsize{11}{60}\selectfont 48.99 & \fontsize{11}{60}\selectfont 15.06 & \fontsize{11}{60}\selectfont 20.13 & \fontsize{11}{60}\selectfont 28.06 \\

\fontsize{10.5}{60}\selectfont \textsc{BertSumExtMulti~\shortcite{Sotudeh2020OnGE}} & \fontsize{11}{60}\selectfont \textbf{53.11} & \fontsize{11}{60}\selectfont 16.77 & \fontsize{11}{60}\selectfont 20.34 & \fontsize{11}{60}\selectfont  \textbf{30.07} \\

\bottomrule
\end{tabular}
}
\caption{\textsc{Rouge (F1)} results of different systems on the blind test set of {LongSumm} dataset containing 22 abstractive summaries.}
\label{tab:blind}

\end{center}
\end{table}

\subsection{Experimental evaluation} 
The recent effort in extended summarization and its shared task of {LongSumm}~\cite{Chandrasekaran2020LongSumm} used average \textsc{Rouge (F1)} to rank the participating systems, in addition to commonly-used \textsc{Rouge-n} scores. \textcolor{black}{Table \ref{tab:blind} shows the performance of the participated systems on the blind test set. 
As shown, \textsc{BertSumExtMulti} model outperforms other models by a large margin (i.e., with relative improvements of 6\% and 3\% on \textsc{Rouge-1} and average \textsc{Rouge(F1)}, respectively); hence, we use the best-performing in terms of F1 (i.e.,  \textsc{BertSumExtMulti} model) in our experiments. }
Tables. \ref{tab:main} presents our results on the test sets of arXiv-Long and PubMed-Long datasets, respectively. 
 As observed, our model statistically significantly outperforms the state-of-the-art systems on both datasets across most of the \textsc{Rouge} variants, except \textsc{Rouge-L} on PubMed-Long. The improvements gained by our model validates our hypothesis that incorporating the salient introductory sentence representations into the extractive summarizer yields a promising improvement.
Two non-neural models (i.e., \textsc{LSA} and \textsc{LexRank}) underperform the neural models, as expected. 
Comparing the abstractive model (i.e., \bart) with extractive neural ones (i.e., \textsc{BertSumExt} and \textsc{BertSumExtMulti}), we see that while there is relatively a smaller gap in terms of \textsc{Rouge-1}, the gap is larger for \textsc{Rouge-2}, and \textsc{Rouge-L}. Interestingly, in the case of \textsc{Bart}, we found that generating extended summaries is rather challenging for abstractive summarizers. 
Current abstractive summarizers including \bart{} have difficulty in abstracting very detailed information, such as numbers, and quantities, which hurts the faithfulness of the generated summaries to the source. This behavior has a detrimental effect, specifically, on \textsc{Rouge-2} and \textsc{Rouge-L} as their high correlation with human judgments in terms of faithfulness has been shown \cite{Pagnoni2021UnderstandingFI}. 
Comparing the extractive \textsc{BertSumExt} and \textsc{BertSumExtMulti} models, while \textsc{BertSumMultiExt} is expected to outperfom \textsc{BertSumExt}, it is observed that they perform almost similarly, with small (i.e., insignificant) improved metrics. 
This might be due to the fact that \textsc{BertSumExtMulti} works out-of-the-box when a handful amount of sentences are sampled from diverse sections to form the oracle summary as also reported by its authors. However, when labeling oracle sentences in our framework (i.e., Intro-guided labeling), there is no guarantee that the final set of oracle sentences are labeled from diverse sections. Overall, our model achieves about 1.4\%, 2.4\%, 3.5\% (arXiv-Long), and 1.0\%, 2.5\%, 1.3\% (PubMed-Long) improvements across \textsc{Rouge} score variants; and 2.2\% (arXiv-Long), 1.4\% (PubMed-Long) improvements over F1, compared to the neural baselines (i.e., \textsc{BertSumExt} and \textsc{BertSumExtMulti}). \textcolor{black}{While comparing our model with \textsc{BertSumExt-Intro}, we see the vital effect of adding second encoder at finding supplementary sentences across non-introductory sections, where our model gains relative improvements of 9.62\%-26.26\%-16.09\% and 9.40\%-5.27\%-9.99\% for \textsc{Rouge-1, Rouge-2, Rouge-L} on arXiv-Long and PubMed-Long, respectively. In fact, { the sentences that are picked as summary from the introduction section are not comprehensive as such they are \textit{clues} to the \textit{main points} of the paper. The other important sentences are picked from the supplementary parts (i.e., non-introductory) of the paper.} }

\begin{table*}[t]
    \centering
    \begin{tabular}{cc}
         \scalebox{0.9}{
    \begin{tabular}{l@{\hskip 2.5em}rrrr}
         \toprule
         Metric &  Win & Tie & Lose  & agr. \\
         \midrule
         \multicolumn{5}{c}{\small Our Model vs. \textsc{BertSumExt} baseline} \\
         \midrule
         Cohesion &   43\% & 25\% & 32\% & 46.5\%\\
         Completeness & 46\% & 34\% & 20\%  & 48.9\% \\
         \midrule
         \multicolumn{5}{c}{\small Our Model vs. \textsc{BertSumExtMulti} baseline} \\
         \midrule
         Cohesion &  42\% & 24\% & 34\% & 47.2\% \\
         Completeness  & 45\% & 32\% & 24\% & 49.1\%\\
         \bottomrule
         
    \end{tabular}
    } & 
         
         \scalebox{0.9}{
    \begin{tabular}{l@{\hskip 2.5em}rrrrr}
         \toprule
        
         Metric &  Win & Tie & Lose & agr.  \\
         \midrule
         \multicolumn{5}{c}{\small Our Model vs. \textsc{BertSumExt} baseline} \\
         \midrule
         Cohesion &    39\% & 21\% & 30\% & 52.1\% \\
         Completeness &  47\% & 19\% & 34\% & 51.3\%\\
         \midrule
         \multicolumn{5}{c}{\small Our Model vs. \textsc{BertSumExtMulti} baseline} \\
         \midrule
         Cohesion &  37\% & 21\% & 32\% & 48.2\%\\
         Completeness  & 41\% & 17\% & 32\% & 46.3\% \\
         \bottomrule
         
    \end{tabular}
    }
    
    \vspace{0.1cm}
    \\
    \vspace{0.1cm}
     (a)  & (b)


    \end{tabular}
     \vspace{-1em}
    \caption{Results of human evaluations over 40 papers sampled from (a) arXiv-Long's, and (b) PubMed-Long's test set. agr. shows inter-rater agreement. }
    
    \label{tab:eval1}
\end{table*}

\subsection{Human evaluation}
While our model statistically significantly improves upon the state-of-the-art baselines in terms of \textsc{Rouge} scores, a few works have reported the low correlation of \textsc{Rouge} with human judgments~\cite{Liu2008CorrelationBR, Cohan2016RevisitingSE, Fabbri2021SummEvalRS}. In order to provide insights into why and how our model outperforms the best-performing baselines, we perform a manual analysis of our system's generated summaries, \textsc{BertSumExt}'s, and \textsc{BertSumExtMulti}'s. 
For the sake of evaluation, two annotators were asked to manually evaluate two sets of 40 papers' ground-truth abstracts (40 for arXiv-Long, and 40 for PubMed-Long) with their generated extended summaries (baselines' and ours) to gain insights into qualities of each model. Annotators were Electrical Engineering and Computer Science PhD students and familiar with principles of reading scientific papers. Samples were randomly selected from the test set, one from each 40 evenly-spaced bins sorted by the difference of \textsc{Rouge-L} between two experimented systems. 

The evaluations were performed according to two metrics: (1) \textit{Cohesion}: whether the ordering of sentences in summary is cohesive, namely sentences entail each other. (2) \textit{Completeness}: whether the summary covers all salient information provided in the ground-truth summary. To prevent bias in selecting summaries, the ordering of system-generated summaries were shuffled such that it could not be guessed by the annotators. Annotators were asked to specify if the first system-generated summary wins/loses or ties with the second system-generated summary in terms of qualitative metrics. \textcolor{black}{It has to be mentioned that since our model is purely extractive, it does not introduce any fact that is unfaithful to the source.}

    

          
          
          

\begin{figure*}
    \centering
    
    \begin{tabular}{cc}
        \begin{tabular}{c}
             \includegraphics[scale=0.51]{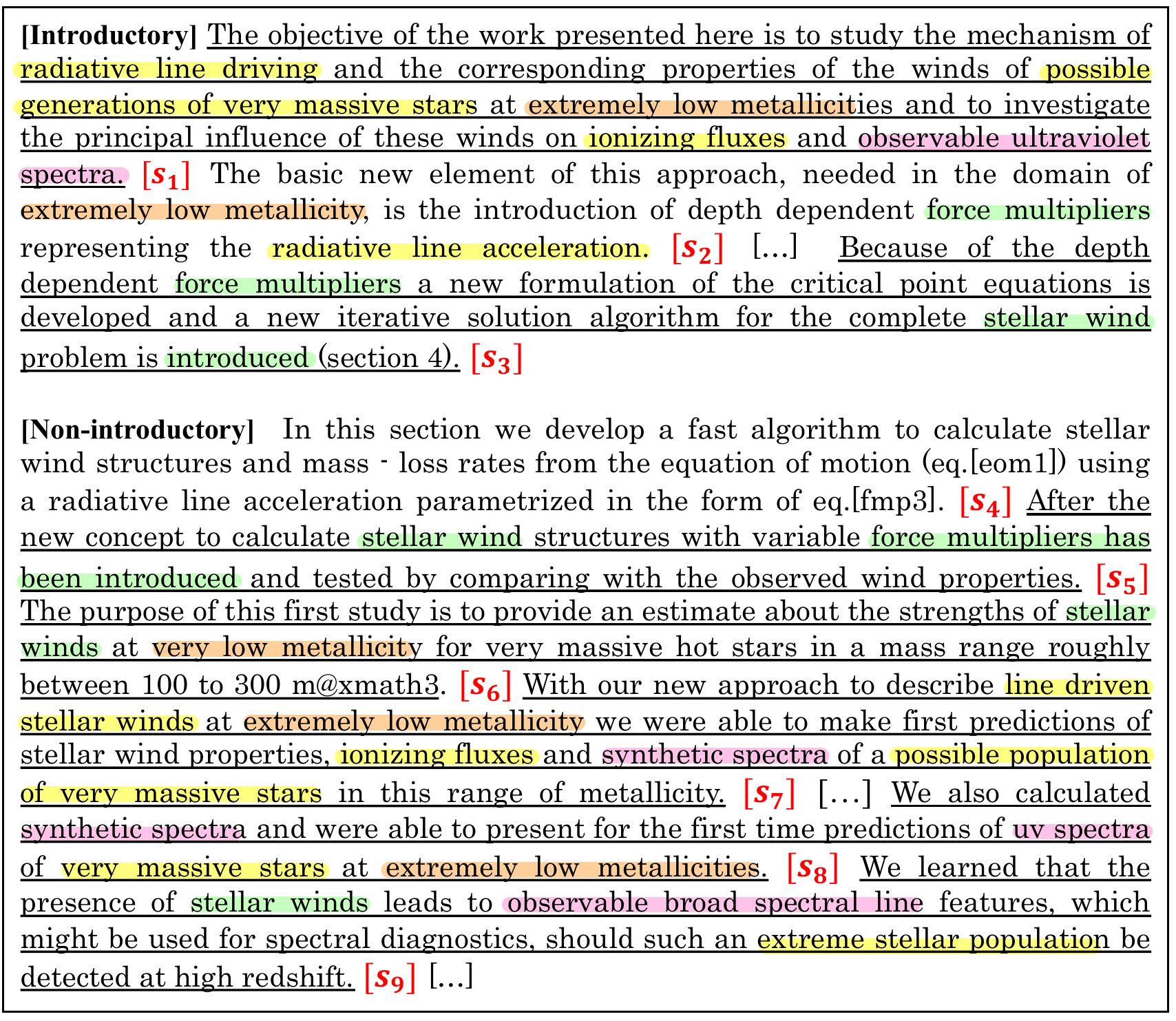} 
        \end{tabular}
        &  
        \begin{tabular}{c}
         \includegraphics[scale=0.14]{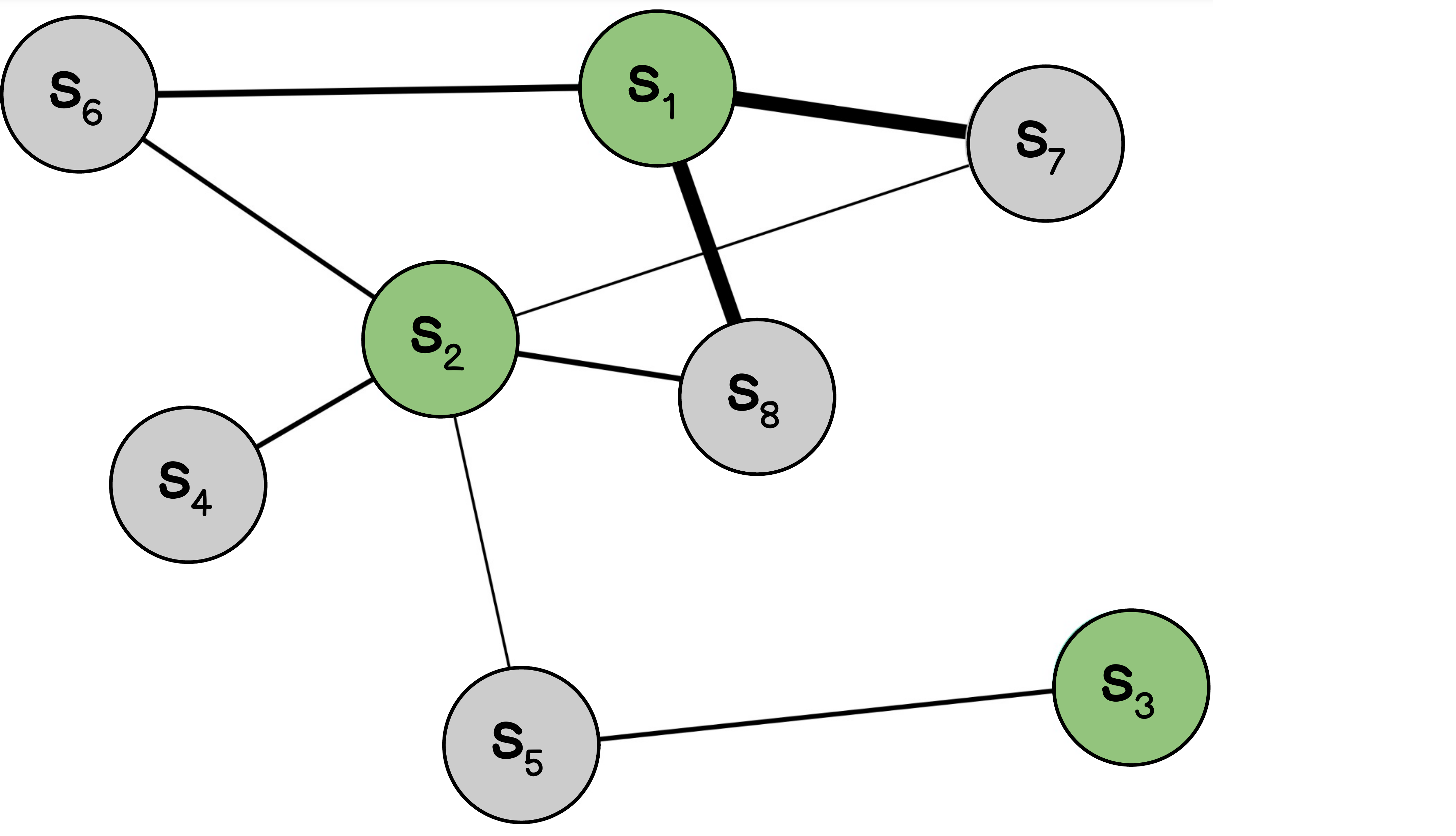}
        \end{tabular}

        \\
         (a)
         & 
         (b) \\
    \end{tabular}
\vspace{-0.5em}

    \caption{(a) Our system's generated summary, (b) Sentence graph visualization of our system's generated summary. Green and gray nodes are introductory and non-introductory sentences, respectively. Edge thickness denotes the \textsc{Rouge} score strength between pair of sentences. Parts, from which sentences are sampled, are shown inside brackets. The summary is truncated due to space limitations. Ground-truth summary-worthy sentences are underlined, and colored spans show pointers from introductory to non-introductory sentences.}
    \label{fig:an_samples}
\end{figure*}
Our human evaluation results along with Cohen's kappa~\cite{Cohen1960ACO} inter-rater agreements are shown in Table \ref{tab:eval1} (agr. column). As shown, our system's generated summaries improve completeness and cohesion in over 40\% for most of the cases (6 out of 8 for win cases~\footnote{Win cases are the ones in which our system wins the baseline(s) in terms of cohesion/completeness.}). Specifically, when comparing with \textsc{BertSumExt}, we see that 68\%, 80\% (arXiv-Long); and 60\%, 66\% (PubMed-Long) of sampled summaries are at least as good as or better than the corresponding baseline's generated summaries in terms of cohesion and completeness, respectively. Overall, across two metrics for \textsc{BertSumExt} and \textsc{BertSumExtMulti}, we gain relative improvements over the baselines: 25.6\%, 19.0\% (cohesion), and 56.5\%, 46.7\% (completeness) on arXiv-Long; and 23.1\%, 13.5\% (cohesion), and 27.7\%, 21.9\% (completeness) on PubMed-Long.~\footnote{Relative improvement of win rate over lose rate.} These improvements, qualitatively evaluated by the human annotators, show the promising capability of our purposed model in generating improved extended summaries which are more preferable than the baselines'. 
We observe a similar improvement trend when comparing our summaries with \textsc{BertSumExtMulti}, where 66\%, 77\% (arXiv-Long); and 58\%, 58\% (PubMed-Long) of our summaries are as good as or better than the baseline's in terms of cohesion and completeness.
Looking at the Cohen's inter-rater agreement, the correlation scores fall into ``moderate'' agreement range according to the interpretation of Cohen's kappa range~\cite{McHugh2012InterraterRT}. 

\subsection{Case study}
Figure \ref{fig:an_samples} (a) demonstrates an extended summary generated from a sample arXiv-Long paper by our model. The underlined sentences denote that the corresponding sentences are oracle (i.e., summary-worthy), the colored spans denote the pointers from introductory information to non-introductory information, and sentence numbers appear in brackets following each sentence. As shown, our system first identifies salient introductory sentences (i.e., $[s_1]$ and $[s_3]$), and then augments them with important non-introductory sentences. Figure \ref{fig:an_samples} (b) shows the \textsc{Rouge} scores between pairs of introductory and non-introductory sentences. The edge thickness signifies the strength of the \textsc{Rouge} score between a pair of sentences. 
For example, introductory sentence $[s_1]$ highly correlates with non-introductory sentence $[s_7]$ as it has a stronger edge ($s_1$, $s_7$) thickness. More specifically, $[s_1]$ has mentions of 
\emph{``radiative line driving''}, 
\emph{``properties of the winds''},
\emph{``possible generations of very massive stars''}, 
and \emph{``ionizing fluxes''} 
which maps to $[s_7]$ with semantically similar mentions of 
\emph{``line driven stellar winds''}, 
\emph{``stellar wind properties''}, 
\emph{``possible generations of very massive stars''}, 
and \emph{``ionizing fluxes''}~\footnote{The entire system-generated summaries are publicly available at \url{https://github.com/Georgetown-IR-Lab/TSTRSum}, including 40 human-evaluated cases.}.

\section{Error Analysis}
To determine the limitations of our model, we further analyze our system's generated summaries and report three common defects, along with the percentage of these errors among underperformed cases. We found that (1) our end-to-end system's performance is highly dependent on the introductory sentence extraction task's performance (i.e., task $\mathbf{t_2}$ in Figure \ref{fig:model}) as identification of salient introductory sentences (i.e., oracle introductory sentences) sets up a firm ground to explore detailed sentences from the non-introductory parts of the paper. In other words, identification of non-salient introductory sentences leads to a drift in finding supplemental sentences from the non-introductory parts. Our model often underperforms when it cannot find important sentences from the introductory part (65\%); (2) in underperformed cases, our model fails in selecting motivation, objective sentences from the introductory part, and only identifies the contribution sentences (i.e., describing paper's contributions), such that the final generated summary is composed of contribution sentences, rather than objective sentences. This observation hurts the system in cohesion and completeness (40\%); and (3) as discussed, our model matches introductory sentences with sentences from non-introductory parts of the paper. Given that two sentences within a scientific paper might conceptually convey the exact same information, but are just paraphrased of each other, our model samples both to form the final summary as a high semantic correlation exists between them. This phenomenon leads to sampling two sentences that convey the same information without providing more details; hence, information redundancy (35\%).


\section{Conclusion}
In this work, we propose a novel approach to tackle the extended summary generation for scientific documents. Our model is built upon the fine-tuned contextualized language models for text summarization. Our method improves over strong and state-of-the-art summarization baselines by adding an auxiliary learning component for identifying salient introductory information of long documents, which are then used as pointers to guide the summarizer to pick summary-worthy sentences. The extensive intrinsic and human evaluations show the efficacy of our model in comparison with the state-of-the-art baselines, using two large scale extended summarization datasets . Our error analysis further paves the path for future reseacrh.

\bibliography{anthology,custom}

\appendix

\label{sec:appendix}

\end{document}